\documentclass[conference]{IEEEtran}
\IEEEoverridecommandlockouts
\usepackage{cite}
\usepackage{amsmath,amssymb,amsfonts}
\usepackage{algorithmic}
\usepackage{graphicx}
\usepackage{textcomp}
\usepackage{xcolor}
\usepackage{float}
\usepackage{stfloats}
\usepackage{booktabs}
\usepackage{hyperref}

\def\BibTeX{{\rm B\kern-.05em{\sc i\kern-.025em b}\kern-.08em
    T\kern-.1667em\lower.7ex\hbox{E}\kern-.125emX}}
\begin{document}
\pagestyle{plain}

\title{Zero-Shot Monocular Depth Estimation for Visual Perception in Lower-Limb Prostheses\\
\thanks{The authors gratefully acknowledge the computing resources provided by the NHR@FAU HPC-Portal at Friedrich-Alexander-Universität Erlangen-Nürnberg.}
}

\author{\IEEEauthorblockN{Ishi Vijayvergiya, Marc Hölle, Youssef Dawoud, Amir El-Ghoussani, Michele De Vita, Rohan Asthana, \\
Philipp Beckerle, and Vasileios Belagiannis}
\IEEEauthorblockA{\textit{Friedrich-Alexander-Universität Erlangen-Nürnberg}\\
Erlangen, Germany \\
\{ishi.vijayvergiya, marc.hoelle, youssef.dawoud, amir.el-ghoussani,\\
michele.de.vita, rohan.asthana, philipp.beckerle, vasileios.belagiannis\}@fau.de}
}

\maketitle
\thispagestyle{plain}

\begin{abstract}
Monocular depth estimation has advanced rapidly with the development of foundation models that demonstrate strong zero-shot generalisation across diverse visual domains. However, the applicability of these models to visual perception for lower-limb prosthetics remains largely unexplored. Reliable depth perception is particularly important for prosthesis users, who often experience significant challenges when navigating stairs, uneven terrain, and other complex environments. To address this gap, we introduce the Lower-Limb Prosthesis (LP) dataset, a benchmark designed to evaluate depth perception from an emulated knee-level prosthetic viewpoint. Using this dataset, we evaluate four state-of-the-art monocular depth estimation foundation models. Our results provide the first systematic benchmark of monocular depth estimation performance from a prosthetic viewpoint, while highlighting both the challenges and the potential of deploying foundation models in prosthetic perception systems. The code and dataset are available at \url{https://github.com/Ishivijay/lp_dataset}.
\end{abstract}


\section{Introduction}
Visual perception is fundamental to autonomous navigation and environmental understanding. In the context of medical prosthetics, the ability to accurately perceive and interpret the surrounding environment is critical for safe and effective device operation. However, current lower-limb prostheses lack robust perceptual capabilities, leaving users at significantly increased risk of falls, particularly when navigating stairs, uneven terrain, and cluttered environments~\cite{miller2001prevalence}. Integrating reliable depth perception into prosthetic systems could enable real-time hazard detection, adaptive gait planning, and enhanced environmental awareness. By improving the ability to navigate complex environments safely, such capabilities could reduce fall risk, enhance mobility, and increase user confidence in daily activities.

\begin{figure}
    \centering
    \includegraphics[width=0.45\linewidth]{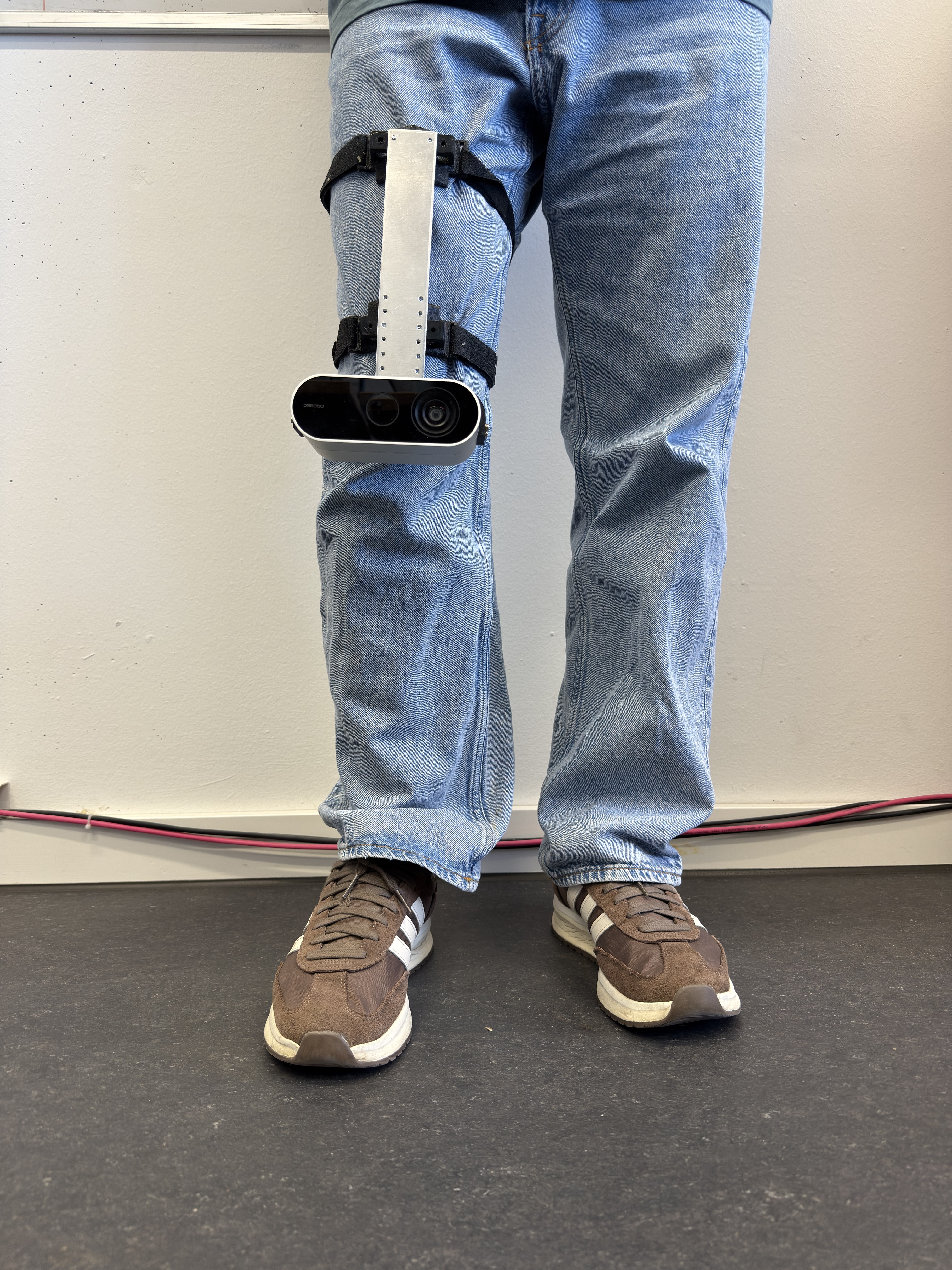}
    \includegraphics[width=0.45\linewidth]{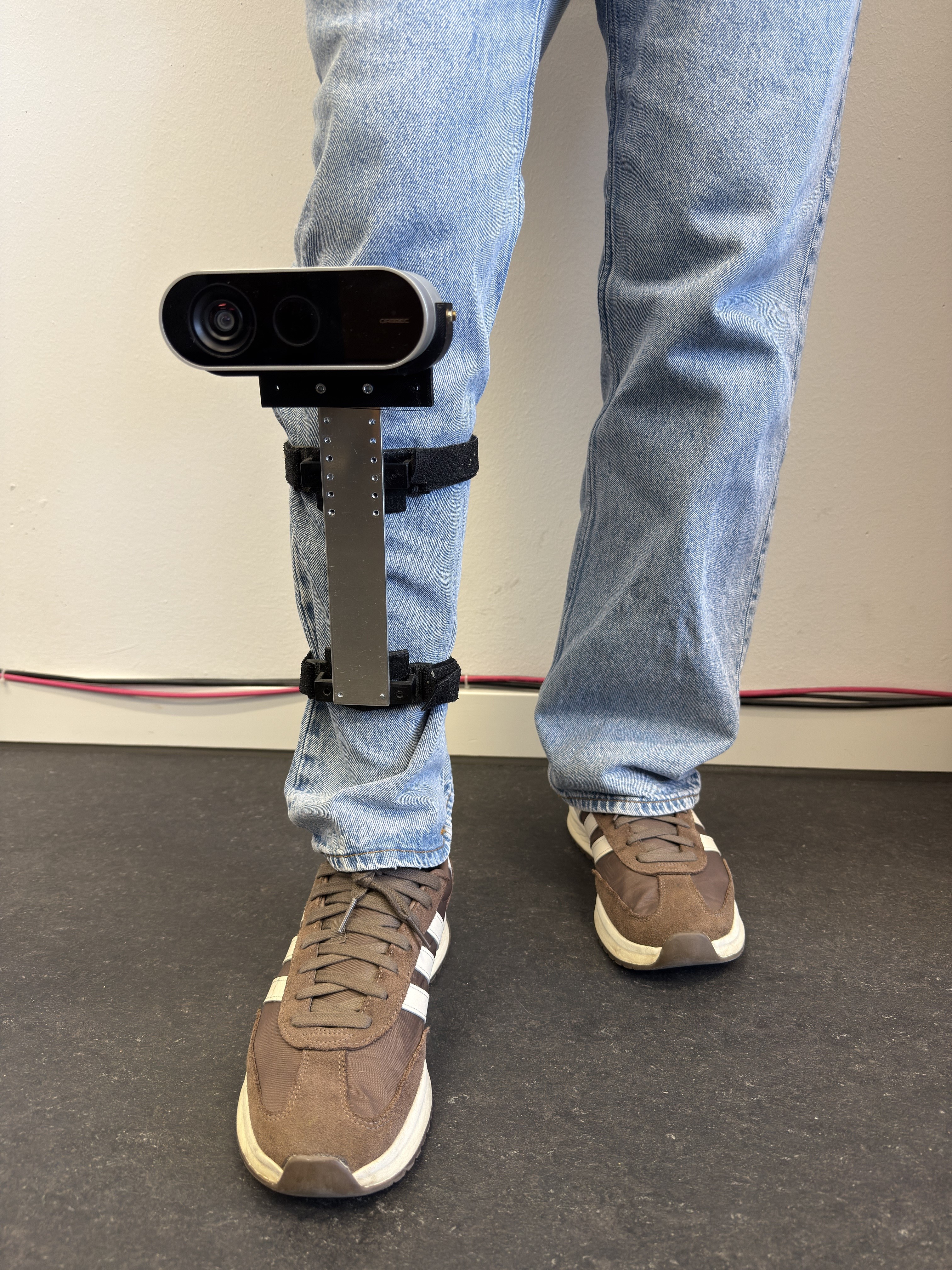}
    \caption{\textbf{LP dataset acquisition setup.} The dataset was collected with a wearable camera mounted on the lower limb in four configurations: above or below the knee, on each leg. These configurations yield egocentric recordings from an emulated knee-level prosthetic viewpoint during locomotion across a diverse set of mounting positions.}
    \label{fig:setup}
\end{figure}

Computer vision approaches are increasingly being integrated into lower-limb wearable robotics to provide active environmental awareness. In lower-limb prosthetics, vision-in-the-loop control using gaze-tracking cues can be used to identify user intention and anticipate terrain transitions, providing volitional control of active knee-ankle prostheses~\cite{chen2025gaze}. Visual-inertial odometry can be combined with 3D terrain reconstruction to classify locomotion~\cite{zhao2023learning}. Similarly, depth-aware pose estimation has also been integrated into lower-limb exoskeleton systems to enable real-time gait analysis and prevent asymmetric locomotion and falls~\cite{wang2023depth}, as well as to facilitate gesture-driven control interfaces leveraging monocular depth features~\cite{qiu2025vision}. While these approaches demonstrate the value of vision, they generally rely on task-specific training, custom multicamera setups, or cooperative rehabilitation environments.

Monocular depth estimation has become an important capability in the field of computer vision~\cite{laina2016deeper}. Recent foundation models have achieved impressive zero-shot generalisation across a variety of scenes. Models such as Marigold~\cite{ke2024repurposing}, Depth Anything V2~\cite{yang2024depth}, Depth Pro~\cite{bochkovskiy2025depth}, and VGGT~\cite{wang2025vggt} employ a range of training strategies to predict dense depth maps from single RGB images. However, these models are typically developed and benchmarked on conventional datasets featuring a standard eye-level perspective.  These advances raise a natural question:  \emph{Can state-of-the-art monocular depth estimation foundation models generalize zero-shot to the unique, highly dynamic, and knee-level perspective of a prosthesis-mounted camera?}

While prior datasets have captured egocentric or wearable camera images of locomotion environments, they focus mainly on RGB-based scene classification or activity recognition and do not provide paired ground-truth depth annotations. Existing depth estimation benchmarks, on the other hand, target indoor scenes NYUv2~\cite{silberman2012indoor}, ScanNet~\cite{dai2017scannet}, or autonomous driving KITTI~\cite{geiger2012we}, none of which reflect the specific viewing angle, motion patterns, and close-range stair navigation experienced by a prosthesis user. To our knowledge, no existing dataset provides paired RGB-depth data recorded from a lower-limb-mounted camera during locomotion, capturing the perspective of a lower-limb prosthesis.

To address this gap, we introduce the Lower-limb Prosthesis (LP) dataset, the first dataset designed specifically for visual perception in medical prosthesis applications. The LP dataset consists of synchronized RGB-depth image pairs captured from a camera mounted on a lower-limb during stair ascent and descent across six recording sessions. The dataset covers a variety of scene elements, including staircases, corridors, landings, and doorways, all captured from an emulated knee-level prosthetic viewpoint during locomotion.

Using this dataset, we carry out a comprehensive zero-shot evaluation of four state-of-the-art monocular depth estimation foundation models. This study provides the first empirical evidence on the robustness and limitations of modern depth estimation models when applied to data from a lower-limb-mounted camera emulating a prosthetic viewpoint, enabling novel research for visual perception in medical prostheses.

The main contributions of this paper are:
\begin{enumerate}
    \item We introduce the LP dataset, the first dataset for visual depth perception for lower-limb prosthesis, comprising six stair-navigation sequences with ground-truth depth from a depth camera.
    \item We present a systematic zero-shot evaluation of four state-of-the-art monocular depth estimation foundation models (Marigold, Depth Anything V2, Depth Pro, and VGGT) on this novel  domain.
    \item We provide a detailed analysis of model performance across different scene types and discuss the implications for integrating depth estimation into prosthetic perception systems.
\end{enumerate}

\section{Related Work}

\subsection{Monocular Depth Estimation}
Monocular depth estimation aims to infer scene depth from a single RGB image and has been extensively studied in computer vision. Early supervised approaches relied on paired RGB–depth datasets, such as NYU Depth V2~\cite{silberman2012indoor} and KITTI~\cite{geiger2012we}, and employed convolutional neural networks for depth prediction~\cite{eigen2014depth}. To reduce dependence on costly ground-truth depth annotations, self-supervised methods were subsequently introduced, leveraging photometric consistency across video frames or stereo image pairs as a supervisory signal~\cite{godard2019digging}. More recently, research has shifted toward foundation models designed to generalize across diverse domains and visual environments. Depth Anything~\cite{yang2024depthDA} and its successor, Depth Anything V2~\cite{yang2024depth}, are trained at scale using a combination of labelled and unlabelled data. Marigold~\cite{ke2024repurposing} repurposes the Stable Diffusion architecture for monocular depth estimation through a diffusion-based denoising framework. Depth Pro~\cite{bochkovskiy2025depth} combines a DINOv2 backbone with a dedicated focal-length estimation module, while VGGT~\cite{wang2025vggt} leverages vision transformers to achieve strong depth estimation accuracy and cross-domain generalisation. Despite their impressive performance on standard benchmark datasets, the robustness of these models under the unconventional camera viewpoints encountered in lower-limb prosthetic applications remains largely unexplored.

\subsection{Egocentric and Wearable Vision Datasets}
Several datasets have been developed using wearable or body-mounted cameras to support mobility and visual perception research. For example, the ExoNet database~\cite{laschowski2020exonet} provides wearable-camera images of walkable environments for scene classification, while other datasets have been proposed for depth estimation using head-mounted or chest-mounted cameras. However, these datasets are primarily designed for general egocentric perception and do not capture the unique visual characteristics of a lower-limb prosthesis-mounted camera, including its low viewpoint, reduced camera height, and distinctive motion dynamics during locomotion and stair negotiation. As a result, they provide limited insight into the performance of depth estimation models in prosthetic navigation scenarios, highlighting the need for datasets specifically designed to evaluate perception from a prosthetic perspective.

\subsection{Visual Perception for Assistive Devices}
Research on intelligent lower-limb prostheses has traditionally focused on mechanical design, control strategies, and electromyography (EMG)-based intent recognition~\cite{tucker2015control}. Recent work started incorporating environmental sensing to improve prosthetic functionality. Examples include the use of wearable-camera imagery for locomotion environment classification~\cite{laschowski2020exonet} and the identification of walkable surfaces from egocentric images~\cite{laschowski2022environment}. While these approaches provide valuable semantic information about the environment, they do not explicitly capture its geometric structure. Incorporating depth-aware perception into prosthetic systems could enable advanced capabilities such as obstacle detection, step-height estimation, terrain assessment, and real-time path planning. Despite these potential benefits, the application of monocular depth estimation to lower-limb prosthetic navigation remains largely unexplored.

\section{Dataset}

\subsection{Dataset Acquisition and Setup}
The LP dataset comprises indoor scenes collected in a multi-storey building using an RGB camera and a co-located depth camera mounted on a participant’s knee to emulate the viewpoint of a knee-level prosthesis-mounted camera. Representative examples are shown in Fig.~\ref{fig:qualitative_comparison}. Data were collected under controlled conditions in a relatively uncluttered indoor setting to establish a clean baseline benchmark and avoid recording unconsented individuals; door thresholds, densely furnished spaces, other people, and moving animals were excluded. This initial collection focused on constant-speed walking in the selected indoor setting. In total three healthy male participants in their mid-twenties took part in the study. One participant was from Europe, and the other two were from Asia. The height of each individual varied from 1.7 m to 1.87 m. All measurements were collected while participants walked at a constant speed. Data were acquired using four mounting configurations: left-above (LA), left-below (LB), right-above (RA), and right-below (RB), where the camera was positioned either above or below the left or right knee joint, as shown in Fig.~\ref{fig:setup}. The sensor was secured to the participant's leg using a mounting strap, with power and data transmitted via a USB Type-C 3.2 connection to a laptop carried by an accompanying person. The dataset contains 3,144 synchronized RGB–depth image pairs distributed across six recording sequences encompassing stair ascent, stair descent, and corridor walking, as summarized in Table~\ref{tab:dataset_sequences}. Data acquisition was performed using an Orbbec Femto Bolt depth camera, which operates on Microsoft's Azure Kinect Time-of-Flight (ToF) technology. The device features a 1-megapixel indirect ToF (iToF) depth sensor configured in wide field-of-view (WFOV) mode, yielding a 120° × 120° field of view and producing depth maps at a resolution of 1024 × 1024 pixels. Co-registered RGB images were captured via the integrated 4K RGB sensor and synchronized with the depth maps at the hardware level.

\begin{table}[t]
\centering
\caption{Overview of the six recording sequences in the LP dataset.}
\label{tab:dataset_sequences}
\small
\begin{tabular}{lcc}
\toprule
Sequence & Direction & Images \\
\midrule
LA downstairs & Down & 428 \\
LA upstairs   & Up   & 629 \\
LB upstairs   & Up   & 591 \\
RA downstairs & Down & 375 \\
RA upstairs   & Up   & 622 \\
RB downstairs & Down & 499 \\
\midrule
\textbf{Total} & & \textbf{3,144} \\
\bottomrule
\end{tabular}
\end{table}

\subsection{Dataset Structure and Formats}
For each sequence, the dataset provides sharpened RGB images \texttt{sharpen\_rgb}, raw depth maps \texttt{raw\_depth}, colorized depth visualizations \texttt{colored\_depth}, temporally smoothed depth \texttt{smooth\_depth} and transformed depth \texttt{transformed\_depth}.  

\section{Experiments}
We perform zero-shot inference using four state-of-the-art foundation models for monocular depth estimation: Marigold~\cite{ke2024repurposing}, Depth Anything V2~\cite{yang2024depth}, Depth Pro~\cite{bochkovskiy2025depth}, and VGGT~\cite{wang2025vggt}. To ensure a consistent comparison, all models were evaluated using the same three-stage pipeline: (1) zero-shot inference on raw RGB images, (2) post-processing and alignment of predicted depth maps with the spatial dimensions of the ground-truth depth data, and (3) quantitative evaluation using standard depth estimation metrics.

\subsubsection{Model Configurations}
Marigold was evaluated using the pre-trained \texttt{marigold-depth-v1-1} diffusion checkpoint \cite{ke2024repurposing}, with half-precision (FP16) inference enabled. Depth Anything V2 was deployed with the large Vision Transformer encoder (ViT-L) configuration \cite{yang2024depth}, and inference was configured to produce raw depth maps. Depth Pro was loaded using its official model factory, generating metric depth predictions in meters together with estimated camera focal lengths \cite{bochkovskiy2025depth}. VGGT, originally developed for multi-view 3D reconstruction, was adapted for single-image depth estimation using the VGGT-1B checkpoint (1 billion parameters) \cite{wang2025vggt} obtained from Hugging Face. Input images were preprocessed using the official VGGT preprocessing pipeline in crop mode~\cite{wang2025vggt}, and depth maps were extracted from the model outputs for evaluation.

\subsubsection{Evaluation Protocol}
All models were evaluated on the complete set of sequences in the LP dataset under identical experimental conditions. When the spatial resolution of a predicted depth map differed from that of the ground-truth depth map, bilinear interpolation was used to resize the prediction to the ground-truth dimensions.

To enable fair comparison across models with differing scale and shift characteristics, a least-squares alignment procedure was applied per image. Specifically, a global scale factor $s$ and shift term $t$ were estimated by solving
\begin{equation}\label{eqEval}
\text{GT} = s \cdot \text{pred} + t,
\end{equation}
where $\text{GT}$ denotes the ground-truth depth map and $\text{pred}$ denotes the predicted depth map~\cite{ke2024repurposing}. The evaluation metrics included absolute relative error (Abs Rel), squared relative error (Sq Rel), root mean square error (RMSE), root mean square error in logarithmic space (RMSE$_{\log}$), and threshold accuracies $\delta_1$, $\delta_2$, and $\delta_3$. Both the least-squares alignment in Eq.~\ref{eqEval} and all evaluation metrics were computed exclusively over valid ground-truth depth pixels ($GT>0$), excluding unmeasured sensor dropout regions.
%Depth-Pro and VGGT were additionally evaluated without alignment to assess their native metric depth estimation performance. 

\begin{figure*}[h!]
  \centering
  \includegraphics[width=\textwidth]{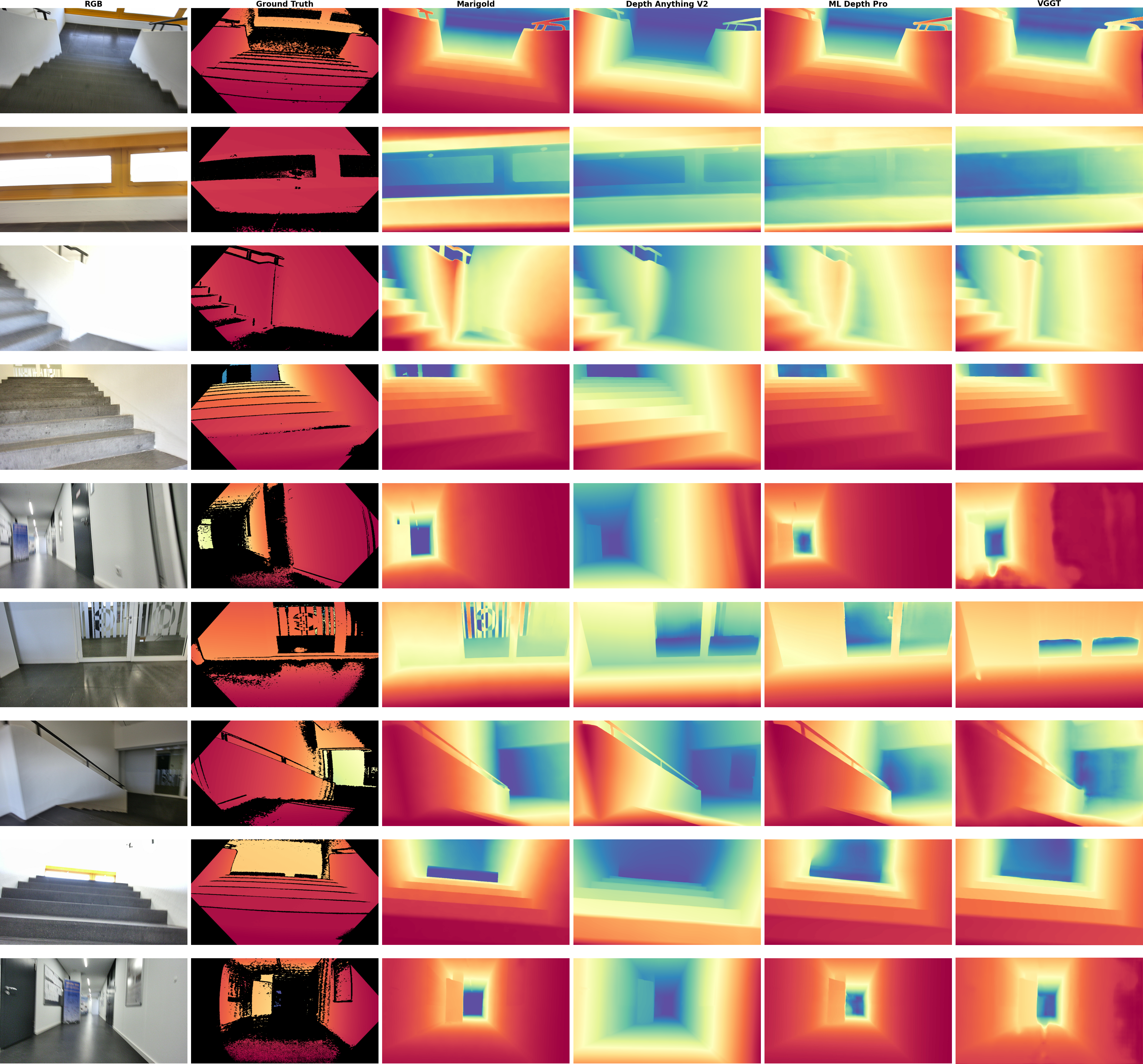}
  \caption{\textbf{Qualitative comparison of zero-shot monocular depth estimation on the LP dataset.} Nine representative scenes (rows), spanning corridors, staircases, landings, and doorways, are shown
with, from left to right, the RGB input, the Ground truth and the depth predicted by Marigold, Depth Anything~V2, Depth Pro, and VGGT. Depth is rendered with a colormap, where warm colors denote near and cool colors denote far surfaces. Black regions in the ground truth denote invalid sensor returns that are caused by surfaces beyond the operating range, grazing-angle specular reflections on floor tiles, or areas outside the depth field of view.}
  \label{fig:qualitative_comparison}
\end{figure*}

\section{Discussion}
\begin{table}[t]
\centering
\caption{Zero-shot depth estimation results on the LP dataset (Staircase scene) with least-squares alignment.
Lower is better for $\downarrow$ metrics; higher is better for $\uparrow$ metrics. Best in \textbf{bold}.}
\label{tab:zero_shot_staircase}
\small
\begin{tabular}{l|cccc}
\toprule
Metric & Marigold & DA-V2 & Depth Pro & VGGT \\
\midrule
Abs Rel $\downarrow$ & 0.1568 & 0.2066 & 0.1599 & \textbf{0.1532} \\
Sq Rel $\downarrow$  & 0.0858 & 0.1158 & 0.0829 & \textbf{0.0818} \\
RMSE $\downarrow$    & 0.3549 & 0.3947 & 0.3489 & \textbf{0.3449} \\
RMSE$_{\log}$ $\downarrow$ & \textbf{0.1972} & 0.2806 & 0.2006 & 0.2011 \\
\midrule
$\delta_1$ $\uparrow$ & 0.8075 & 0.7179 & 0.8008 & \textbf{0.8099} \\
$\delta_2$ $\uparrow$ & \textbf{0.9434} & 0.8955 & 0.9408 & \textbf{0.9434} \\
$\delta_3$ $\uparrow$ & \textbf{0.9756} & 0.9489 & 0.9754 & 0.9755 \\
\bottomrule
\end{tabular}
\end{table}

\begin{table}[t]
\centering
\caption{Zero-shot depth estimation results on the LP dataset (Corridor scene) with least-squares alignment.
Lower is better for $\downarrow$ metrics; higher is better for $\uparrow$ metrics. Best in \textbf{bold}.}
\label{tab:zero_shot_corridor}
\small
\begin{tabular}{l|cccc}
\toprule
Metric & Marigold & DA-V2 & Depth Pro & VGGT \\
\midrule
Abs Rel $\downarrow$ & \textbf{0.1963} & 0.2739 & 0.2219 & 0.2477 \\
Sq Rel $\downarrow$  & \textbf{0.0813} & 0.1513 & 0.0936 & 0.1241 \\
RMSE $\downarrow$    & \textbf{0.3357} & 0.5067 & 0.3595 & 0.4084 \\
RMSE$_{\log}$ $\downarrow$ & \textbf{0.2205} & 0.3980 & 0.2483 & 0.2898 \\
\midrule
$\delta_1$ $\uparrow$ & \textbf{0.7592} & 0.5442 & 0.6632 & 0.6100 \\
$\delta_2$ $\uparrow$ & \textbf{0.9371} & 0.8119 & 0.9184 & 0.8824 \\
$\delta_3$ $\uparrow$ & \textbf{0.9669} & 0.9277 & 0.9635 & 0.9581 \\
\bottomrule
\end{tabular}
\end{table}

\subsection{Quantitative Results}
Tables~\ref{tab:zero_shot_staircase} and~\ref{tab:zero_shot_corridor} summarize the zero-shot depth estimation performance of the four foundation models on the LP dataset following least-squares alignment. On the staircase scene (Table~\ref{tab:zero_shot_staircase}), VGGT achieved the strongest performance, obtaining the best Abs Rel (0.1532), Sq Rel (0.0818), RMSE (0.3449), and $\delta_1$ (0.8099). On the corridor scene (Table~\ref{tab:zero_shot_corridor}), Marigold outperformed all other models across every evaluation metric, achieving the lowest Abs Rel (0.1963), Sq Rel (0.0813), RMSE (0.3357), and RMSE$_{\log}$ (0.2205), together with the highest threshold accuracies ($\delta_1$ = 0.7592, $\delta_2$ = 0.9371, and $\delta_3$ = 0.9669). Depth Pro and VGGT remained competitive across both scenes, whereas Depth Anything V2 consistently produced the largest errors and lowest threshold accuracies. Despite differences in ranking, all models demonstrated effective zero-shot generalisation to the emulated knee-level prosthetic viewpoint, with Marigold and VGGT exhibiting the strongest overall performance.

\subsection{Qualitative Results}
Figure~\ref{fig:qualitative_comparison} presents qualitative comparisons of the depth maps generated by the four foundation models. By visual inspection of Figure~\ref{fig:qualitative_comparison}, all models successfully capture the global scene geometry and depth gradients, clearly distinguishing nearby structures from more distant regions. Depth Anything V2 and VGGT generally produce smoother and more spatially coherent depth maps, preserving large-scale scene layouts across a variety of environments. Marigold also generates plausible depth estimates but appears more sensitive to local structures and depth discontinuities, resulting in finer-grained depth variations. Depth Pro produces metric depth predictions with clear foreground–background separation, although its outputs are occasionally more over-smoothed than those of the other models. Despite the challenging knee-level camera viewpoint and motion characteristics of the lower-limb-mounted setup, all methods generate depth predictions that remain broadly consistent with the underlying scene structure. These qualitative results demonstrate the strong zero-shot generalisation capabilities of contemporary monocular depth estimation foundation models in prosthetic perception scenarios.

\subsection{Future Work}

Although the examined zero-shot foundation models demonstrate strong generalisation, their large parameter sizes and high computational demands currently prevent them from being deployed directly on active prosthetic hardware. To address this limitation, future work should investigate two complementary directions. First, model compression, pruning, and quantization techniques should be explored to enable these large foundation models to run efficiently on resource-constrained hardware. Second, deploying lightweight monocular depth architectures~\cite{HornauerNB21, rudolph2022lightweight} is necessary for real-time edge processing. Real-time gait adaptation requires visual feedback to be processed within the swing phase (typically under 400 ms). Future work should therefore focus on deploying both optimised foundation models and lightweight adapted networks on wearable edge accelerators. Furthermore, because prosthetic locomotion is safety-critical, integrating monocular depth uncertainty estimation~\cite{hornauer2025revisiting} and pixel-wise out-of-distribution detection~\cite{holle2025uncertainty} is essential to identify unreliable depth predictions and unexpected terrain hazards. In this context, the LP dataset provides a relevant benchmark for evaluating the impact of such optimizations on depth estimation performance under prosthesis-specific operating conditions.

\section{Conclusion}
In this study, we present the Lower-Limb Prosthesis (LP) dataset, a benchmark comprising synchronised RGB and ground-truth depth data collected from a knee-level camera mounted on healthy participants to emulate a lower-limb prosthesis. The dataset addresses a significant gap in prosthetic perception research by providing depth-aware visual data captured during stair navigation and indoor locomotion from this emulated prosthetic perspective. Using the LP dataset, we evaluated four state-of-the-art monocular depth estimation foundation models and found that, despite being trained on conventional benchmarks, they generalise effectively to this previously unexplored domain. These baseline results demonstrate the feasibility of using zero-shot foundation models to support downstream prosthetic applications such as obstacle detection, terrain assessment and adaptive gait planning. This opens up the possibility of advanced visual perception in intelligent lower-limb prostheses. By introducing a dedicated benchmark and establishing strong baseline performance, this work provides a foundation for future research into visual perception in lower-limb prostheses.

\bibliographystyle{IEEEtran}
\bibliography{references}

\end{document}